\documentclass{article}

\usepackage{microtype}
\usepackage{graphicx}
\usepackage{subcaption}
\usepackage{booktabs}

\usepackage{hyperref}

\usepackage[preprint]{icml2026}

\usepackage{amsmath}
\usepackage{amssymb}
\usepackage{mathtools}
\usepackage{amsthm}

\usepackage{amsfonts,bm,bbm,array,float,mathdots,dsfont,mathrsfs,latexsym}
\usepackage{caption,multirow,longtable,lscape,wrapfig,tikz,tikz-cd,comment}
\usepackage{xcolor}
\usepackage{url}

\usepackage[capitalize,noabbrev]{cleveref}

\theoremstyle{plain}
\newtheorem{theorem}{Theorem}[section]

\newtheorem{ex}[theorem]{Example}

\theoremstyle{definition}
\newtheorem{definition}[theorem]{Definition}

\theoremstyle{remark}

\newtheorem{rmk}[theorem]{Remark}

\usepackage[textsize=tiny]{todonotes}

\newcommand{\Dyck}{\mathrm{Dyck}}
\newcommand{\area}{\mathrm{area}}
\newcommand{\dinv}{\mathrm{dinv}}
\newcommand{\bounce}{\mathrm{bounce}}

\newcommand{\rev}{\mathrm{rev}}

\newenvironment{red}{\relax\color{red}}{\relax}
\newenvironment{blue}{\relax\color{blue}}{\hspace*{.5ex}\relax}

\newcommand{\ber}{\begin{red}}
\newcommand{\er}{\end{red}}
\newcommand{\beb}{\begin{blue}}
\newcommand{\eb}{\end{blue}}

\usetikzlibrary{calc}

\newcommand\dyckpath[5]{
  \begin{scope}[local bounding box=#4]
    \fill[white] (#1) rectangle +(#2,#2);
    \path[fill,red] (#1) foreach \num [count=\i from 0] in {#5}
      { +(-0.5,\i+0.75) node[anchor=north]{\num} \ifnum\i>#2 circle (1pt) \fi};
    \draw[help lines] (#1) grid +(#2,#2);
    \draw[help lines] (#1)--+(#2,#2);
    \draw[line width=2pt] (#1) foreach \dir in {#3}{ -- ++(\dir*90:1)};
  \end{scope}
}

\icmltitlerunning{Discovering a Zeta Map Algorithm on Dyck Paths via Mechanistic Interpretability}

\begin{document}

\twocolumn[
\icmltitle{Discovering a Zeta Map Algorithm on Dyck Paths via Mechanistic Interpretability}



\begin{icmlauthorlist}
\icmlauthor{Xiaoyu Huang}{temple}
\icmlauthor{Blake Jackson}{icarm}
\icmlauthor{Kyu-Hwan Lee}{uconn,kias}
\end{icmlauthorlist}

\icmlaffiliation{temple}{Department of Mathematics, Temple University, Philadelphia, PA, USA}
\icmlaffiliation{icarm}{Institute for Computer-Aided Reasoning in Mathematics, Carnegie Mellon University, Pittsburgh, PA, USA}
\icmlaffiliation{uconn}{Department of Mathematics, University of Connecticut, Storrs, CT, USA}
\icmlaffiliation{kias}{Korea Institute for Advanced Study, Seoul 02455, Republic of Korea}

\icmlcorrespondingauthor{Xiaoyu Huang}{xiaoyu.huang@temple.edu}


\icmlkeywords{AI for mathematics, human-AI collaboration, AI-assisted discovery, combinatorics, Dyck paths, zeta map, transformers, mechanistic interpretability}

\vskip 0.3in
]

\printAffiliationsAndNotice{}

\begin{abstract}
Machine learning is increasingly used in mathematical discovery, but in mathematics the desired output is often not a prediction itself, but an explicit construction that can be checked independently. We study this setting through the zeta map on Dyck paths, a classical bijection in the combinatorics of the \(q,t\)-Catalan numbers. We train a deliberately small one-layer, one-head encoder--decoder transformer on this map and analyze its learned computation using mechanistic interpretability tools, including decoder cross-attention analysis, linear probing, and causal intervention. The analysis reveals a level-based mechanism: encoder representations make path levels linearly accessible, while the decoder selects and traverses input positions in a structured way. Translating these signals into combinatorics leads to the \emph{scaffolding map}, an explicit peak-centered traversal algorithm for Dyck paths. We prove that this algorithm agrees with the zeta map, modulo a reversal convention in the labeling. This gives a controlled example of AI-assisted mathematical discovery in which mechanistic interpretability turns model behavior into a precise, human-verifiable combinatorial algorithm.
\end{abstract}

\section{Introduction}
Machine learning is increasingly used in mathematical research~\citep{gukov2021learning,charton2024patternboost,novikov2025alphaevolve}, but many successful applications still treat a model primarily as a high-accuracy predictor or generative model.  For mathematical discovery, prediction alone is typically insufficient: the desired output is often a conjecture, construction, or explicit algorithm that a mathematician can verify.  This paper studies such a human-AI workflow in a concrete combinatorial setting, the zeta map on Dyck paths.  We train transformer models~\citep{vaswani2017attention} on a known mathematical bijection, interpret the mechanism learned by a deliberately small transformer via mechanistic interpretability tools, and extract from this interpretation a new explicit algorithm.

The zeta map is a central combinatorial object in the theory of $q,t$-Catalan numbers because it gives a bijective explanation for the symmetry between the standard Dyck-path statistics appearing in the formulas for $q,t$-Catalan numbers. In particular, it converts the $(\dinv,\area)$ description into the $(\area,\bounce)$ description, linking two different definitions~\citep{haglund_conjectured_2003,Haiman2000_qtCatalan}. Since the zeta map already admits several algorithmic descriptions \cite{andrews_ad-nilpotent_2002,haglund_conjectured_2003}, it provides an ideal test case for AI-assisted mathematical discovery: the data are exactly generable, correctness can be checked exhaustively at moderate sizes, and the desired output is a discrete combinatorial procedure rather than a numerical approximation.

We first show that transformers can learn the zeta map across several experimental settings. We then reduce the architecture to a one-layer, one-head encoder--decoder transformer, for which the learned solution is sufficiently structured to analyze mechanistically. The central quantity revealed by this analysis is the level sequence of the input Dyck path: cross-attention suggests that the decoder selects positions according to level, causal ablations show that many upward-step positions are not directly needed during decoding, and linear probes show that level information is recoverable from encoder representations. These observations led us to formulate the \emph{scaffolding map}, a peak-centered traversal algorithm for Dyck paths. Unlike the classical area-sequence descriptions of the zeta map, this procedure is organized directly in terms of levels and local traversal. To the best of our knowledge, this is a first example of using mechanistic interpretation of a trained neural sequence model to extract a new explicit bijective description of a well-studied combinatorial map.

A broader goal of this paper is to use the zeta map as a proving ground for AI in combinatorics research, namely to demonstrate the ability to discover combinatorial bijections and algorithms from trained models. While the precise details may differ from one application of AI in math to another, the concepts articulated in this paper are widely applicable within combinatorics research. The model supplies recurring structural signals; the human investigator interprets these signals as combinatorial operations; additional interventions test whether the interpretation is causally relevant; and the resulting procedure becomes a mathematical object that can be stated and proved independently of the neural network.


\section{Related Work}
\paragraph{AI-Assisted Discovery.} AI methods have shown increasing promise as search tools for mathematical discovery. Reinforcement-learning approaches have been used to find explicit examples and counterexamples in extremal combinatorics, graph theory, and knot theory~\citep{wagner2021constructionscombinatoricsneuralnetworks,gukov2021learning}. 
More recent work further studies how reinforcement-learning agents navigate difficult mathematical search spaces with sparse high-reward instances~\citep{Butbaia2026hierarchical,Fagan2026the}. Complementary generative model-guided search and LLM-based coding agents have also been used to discover extremal mathematical structures and improved algorithms~\citep{charton2024patternboost,berczi2026flow,novikov2025alphaevolve}. 
Our work differs in focus: rather than using AI to search for examples, we use mechanistic interpretability to extract an explicit combinatorial algorithm from a trained model. 

\paragraph{Mechanistic Interpretability.}
Mechanistic interpretability seeks to reverse engineer the algorithms implemented by trained neural networks. 
Relevant precedents in mathematical settings include analyses of small transformers trained on modular arithmetic and group operations~\citep{nanda2023progress,chughtai2023toy,zhong2023clockpizzastoriesmechanistic} and arithmetic and enumerative geometry tasks~\cite{babei2025learning,hashemi2025can}. There are recent efforts on automating discovery such as automated circuit discovery~\citep{conmy2023towards}, automated neuron or feature explanation~\citep{bills2023language,shaham2024multimodal}, and activation-to-language decoding methods~\citep{pan2026latentqateachingllmsdecode,karvonen2026activationoraclestrainingevaluating}. 

\section{Mathematical Background}
\subsection{Dyck paths and $q,t$-Catalan statistics}
\begin{figure}[H]
    \centering
    \includegraphics[width=0.7\linewidth]{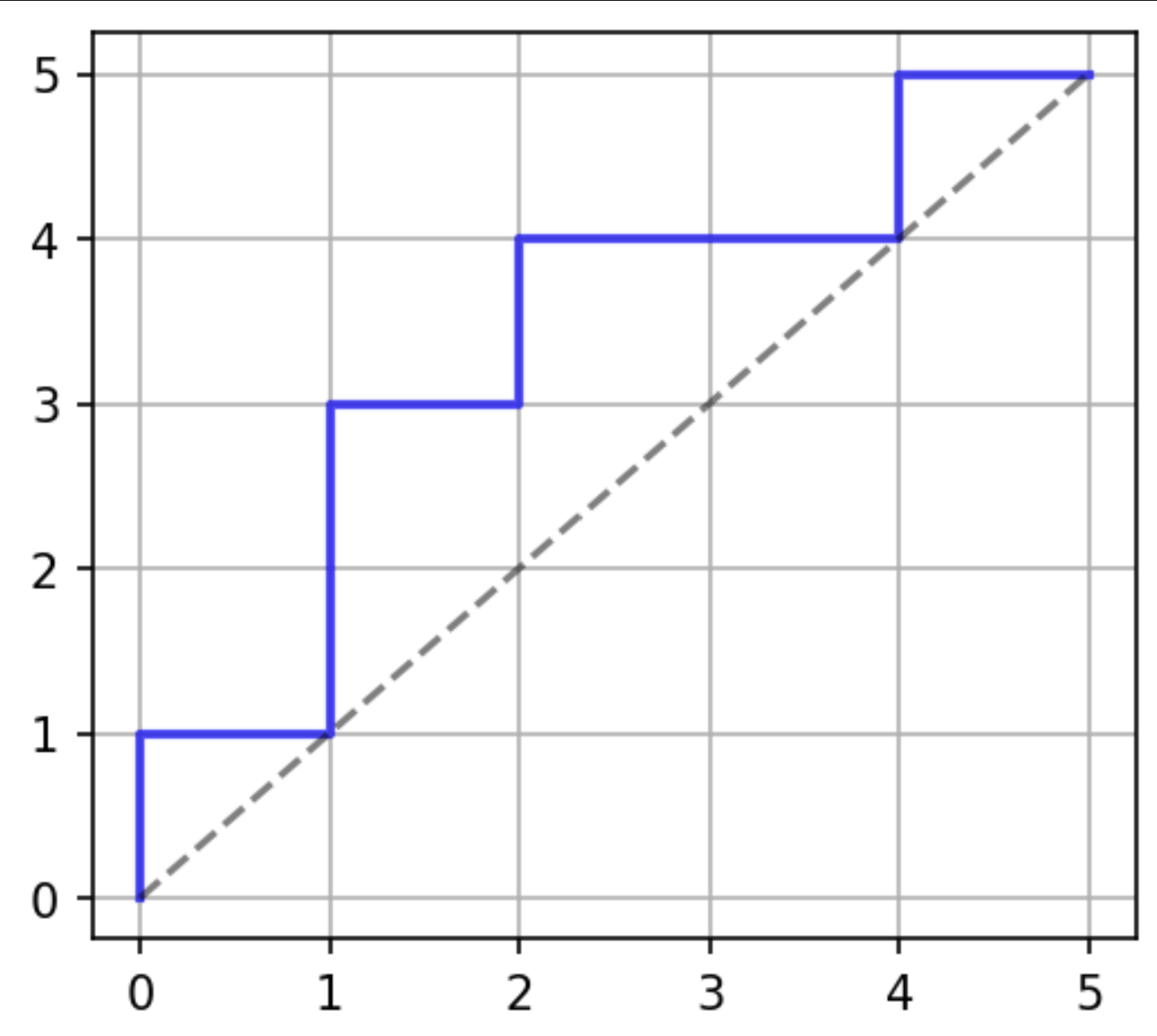}
    \caption{An example of Dyck path of semilength $5$.}
    \label{fig:example}
\end{figure}

Dyck paths are fundamental objects in combinatorics. A \emph{Dyck word} of semilength \(n\) is a binary word \(w=(w_i)_{i=1}^{2n}\) of length \(2n\) consisting of exactly $n$ \(1\)'s such that every initial segment contains at least as many \(1\)'s as \(0\)'s. We write \(\Dyck(n)\) for the set of such words where \(\lvert \Dyck(n)\rvert\) is the well-known \(n\)th Catalan number. Equivalently, by interpreting \(1\) as a North step and \(0\) as an East step, a \emph{Dyck path} is a lattice path from \((0,0)\) to \((n,n)\) that stays weakly above the diagonal \(y=x\), with an example shown in \Cref{fig:example}.

The \(q,t\)-Catalan numbers refine the Catalan numbers and admit two Dyck-path expansions involving the statistics \(\area\), \(\bounce\), and \(\dinv\):
\begin{align}
    C_n(q,t)
        &= \sum_{w\in \Dyck(n)} q^{\area(w)}t^{\bounce(w)} \nonumber \\
        &= \sum_{w\in \Dyck(n)} q^{\dinv(w)}t^{\area(w)}. {\label{eq:two-formulas}}
\end{align}
These formulas were discovered independently by \citep{haglund_conjectured_2003, Haiman2000_qtCatalan}. Here \(\area(w)\) counts the cells between the path and the diagonal, \(\bounce(w)\) is computed from Haglund's bounce path, and \(\dinv(w)\) counts certain diagonal inversions in the area sequence of \(w\). 

The equivalence of the two definitions motivates the search for a bijection that exchanges these statistics. Haglund's zeta map \citep{haglund_conjectured_2003} is a  bijection $\zeta$, satisfying
\[
    (\dinv(w),\area(w)) = (\area(\zeta(w)),\bounce(\zeta(w))).
\]
Although the zeta map is unique, it has several combinatorial interpretations, including descriptions in terms of area sequences
\citep{andrews_ad-nilpotent_2002,haglund_conjectured_2003} and the sweep map~\citep{armstrong2015sweep,thomas2018sweeping}, which describes its inverse.
We do not aim to replace these interpretations. Instead, we ask whether a model trained on the map can reveal an alternative algorithmic organization. We include further background on the \(q,t\)-Catalan numbers and the zeta map in \Cref{app:qt_cat}.

For a Dyck word $w=(w_i)_{i=1}^{2n}$, we define its level sequence by $\ell_0=0$ and
\[
\ell_{i}=\begin{cases}
\ell_{i-1}+1, & \text{ if }w_i=1,\\
\ell_{i-1}-1, & \text{ if }w_i=0.
\end{cases}
\]
Levels later become a central latent combinatorial quantity identified by our interpretability analysis.

\section{Experimental Setup}
\paragraph{Experiments.} For fixed semilength $n$, we construct a supervised dataset of Dyck words and their images.  More precisely, for each input Dyck word \(w\), the target sequence is the image produced by the \textsc{SageMath}~\citep{sagemath} function \texttt{area\_dinv\_to\_bounce\_area\_map()}. Under the labeling convention used by this function, discussed further in \Cref{app:proof}, the target agrees with \(\mathrm{rev}\circ\zeta(w)\), where \(\mathrm{rev}\) denotes reading the Dyck word backward and swapping \(0\)'s and \(1\)'s.

Using a fixed architecture with an embedding dimension of $256$, 4 encoder layers, and 4 decoder layers (each with 8 attention heads), the models achieved $>99\%$ accuracy in learning the zeta map, that is, predicting the image of a Dyck path under the zeta map for datasets of semilengths $n = 11, 12, 13, 14, 15, 16$, with sizes $58{,}786$, $208{,}012$, $742{,}900$, $2{,}674{,}440$, $9{,}694{,}845$, and $35{,}357{,}670$, corresponding to the Catalan numbers $C_n$.

We later varied the architecture and the hyperparameters and explored with a selective subset of
\begin{itemize}
    \item encoder-only, decoder-only, and encoder--decoder architectures;
    \item the number of transformer layers, ranging from 1 to 4;
    \item the number of attention heads, ranging from 1 to 8,
\end{itemize}
and the model was able to obtain near perfect accuracy for almost all settings.

\paragraph{Compression.} We then trained a minimal model with embedding dimension $128$, one encoder layer, one decoder layer, and one attention head. The model has $339{,}716$ parameters and achieved perfect or near-perfect accuracy across repeated runs.  This model trained on the $n = 13$ dataset, which we call the \emph{Minimal Dyck Transformer}, is the focus of the interpretability analysis below. We include architectural details of \emph{Minimal Dyck Transformer} in \Cref{app:minimal_Dyck}.

\section{Interpreting the Minimal Dyck Transformer}
\subsection{Cross-attention patterns}
\begin{figure*}[t]
    \centering
    \includegraphics[width=0.95\textwidth]{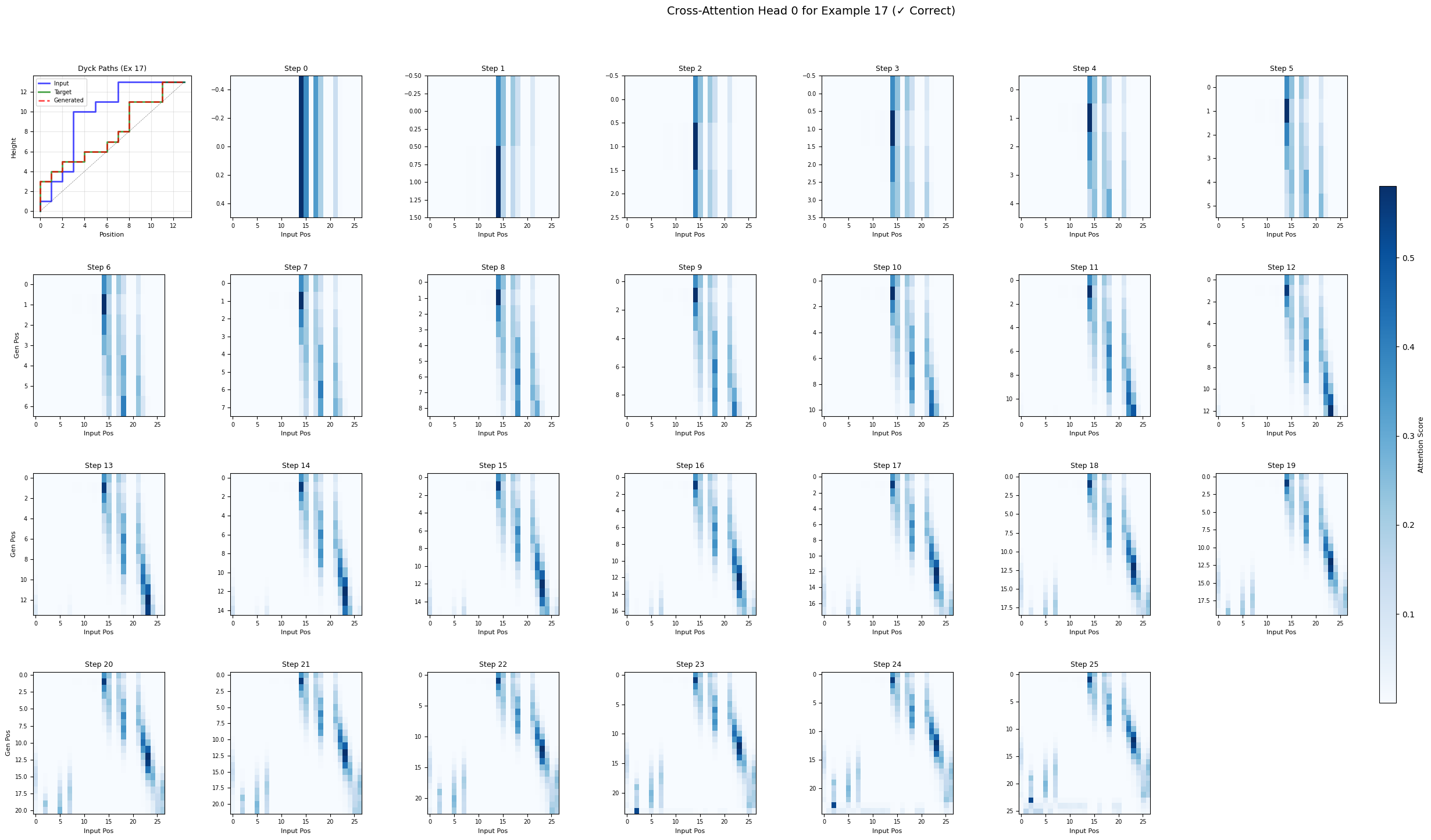}
    \caption{Representative decoder cross-attention matrices for the \textit{Minimal Dyck Transformer}. The recurring sparse patterns indicate that generation is organized by level-based selection rather than by an unstructured lookup table.}
    \label{fig:xatt_example}
\end{figure*}

While several mechanistic inspections reveal useful structure, as shown in \Cref{app:other_visuals}, the most informative internal signal is decoder cross-attention. We dynamically track, at each generation step, which input positions receive the largest decoder cross-attention weights. An example is shown in \Cref{fig:xatt_example}. During generation, the decoder repeatedly attends to a small and structured subset of input positions.  Across examples, we observe three recurring patterns.

\begin{figure*}[t]
    \centering
    \newlength{\panelsize}
    \setlength{\panelsize}{0.32\textwidth}

    \begin{subfigure}[t]{0.32\textwidth}
        \centering
        \includegraphics[
            width=\panelsize,
            height=\panelsize,
            keepaspectratio
        ]{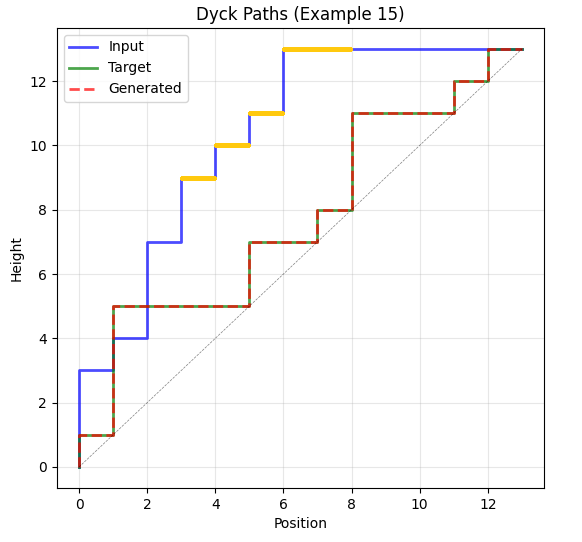}
        \caption{}
        \label{fig:dyck_ex}
    \end{subfigure}
    \hspace{0.01\textwidth}
    \begin{subfigure}[t]{0.32\textwidth}
        \centering
        \includegraphics[
            width=\panelsize,
            height=\panelsize,
            keepaspectratio
        ]{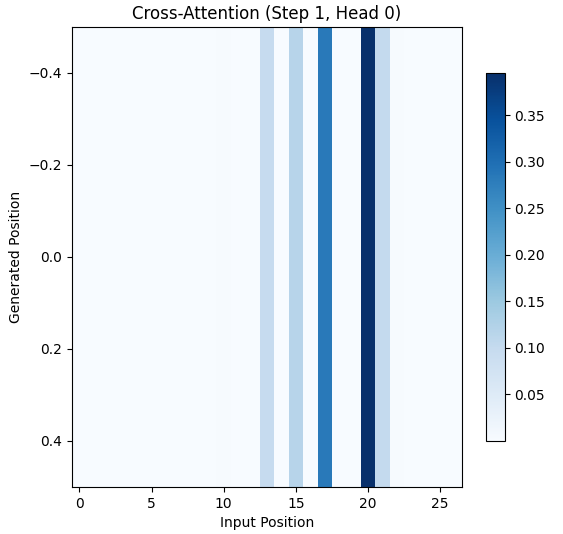}
        \caption{}
        \label{fig:step1_xattn_ex}
    \end{subfigure}
    \hspace{0.01\textwidth}
    \begin{subfigure}[t]{0.32\textwidth}
        \centering
        \includegraphics[
    width=\panelsize,
    height=\panelsize,
    trim={0 0 0 0},
    clip
]{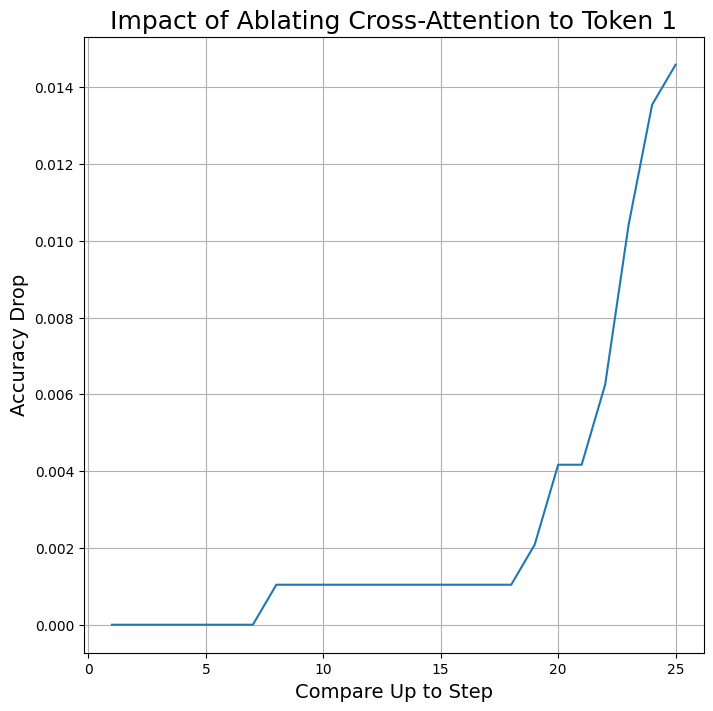}
        \caption{}
        \label{fig:ablation}
    \end{subfigure}

    \vspace{-0.5em}
    \caption{Example of decoder cross-attention and its causal role. Left: input, target, and model-generated Dyck paths, with attended input positions highlighted in yellow. Middle: sparse cross-attention used when generating the first output step on the left. Right: causal ablation results showing negligible decreases in accuracy, measured by comparing the generated output path with the correct target path up to the $k$th step, after ablating all cross-attention to input North steps.}
    \label{fig:attn_ex}
    \vspace{-0.8em}
\end{figure*}

\paragraph{Highest-level selection.}
At early output steps, most prominently observed in the attention visualization when the decoder generates the first step of the output Dyck word, the decoder places attention on input positions whose levels are maximal among relevant positions. An example can be seen in \Cref{fig:dyck_ex} and \ref{fig:step1_xattn_ex}. This suggests that level information has already been computed by the encoder and is being used by the decoder as a selection criterion.

\paragraph{Systematic avoidance of many upward steps.}
Positions with $w_i=1$ receive no cross-attention throughout decoding, hence the decoder cannot directly ``read off’’ level and direction information from upward steps: ignoring $w_i = 1$ prevents the model from directly accessing the upward steps' levels and directions. One way for the decoder to compare information of such positions is that the encoder has already aggregated and stored them in the hidden states of the $w_i = 0$ positions. However, we show this is not the case in \Cref{subsec:ablation/probing}. Instead, it appears to rely heavily on selected positions and reconstructs the rest through a structured traversal.

\paragraph{Triangular sequential shift.}
As decoding proceeds, many examples exhibit the triangular attention pattern shown in \Cref{fig:xatt_example}. This pattern suggests that, after selecting a high-level position $i$ with East-step ($w_i=0$), the model often shifts attention to the next eligible position $i+1$ also with East-step. By the Dyck path structure, this position satisfies $\ell_{i+1}=\ell_i-1$, indicating that the model moves rightward through nearby positions with sequentially decreasing levels.

\subsection{Ablation and probing}
\label{subsec:ablation/probing}
We next test whether these visual patterns correspond to functionally relevant structure.

\paragraph{Causal ablation.}
We mask cross-attention to input positions with $w_i=1$ during decoding. This causal intervention causes only a negligible degradation in accuracy shown in \Cref{fig:ablation}, supporting the hypothesis that many upward-step positions are not essential carriers of decoding-time information.


\paragraph{Linear probing.}
We train linear probes \citep{alain2016understanding} from encoder hidden states to the corresponding level values $\ell_i$.  The probes recover level information with high predictive accuracy of 84\%, indicating that the encoder representation makes this combinatorial statistic linearly accessible.  This supports a two-stage interpretation: the encoder computes level-like information, and the decoder uses it to select and traverse positions.

These observations suggest that the model is not memorizing the zeta map as an arbitrary lookup table. Instead, it appears to organize generation by levels, starting from peaks and then traversing nearby positions. Abstracting this behavior led us to the following explicit procedure, which we call the \emph{scaffolding map}.

\section{The Scaffolding Map}
\label{sec:scaffolding}
We now state the algorithm suggested by the \emph{Minimal Dyck Transformer}.  Let $w=(w_i)_{i=1}^{2n}$ be a Dyck word, and compute levels $\ell_i$ as above.  Let
\[
R=\{i\in\{1,\ldots,2n\}: w_i=0\}
\]
be the set of East-step positions.  A \emph{peak position} is an index $i$ such that $(w_i,w_{i+1})=(1,0)$.

Informally, the algorithm starts from peaks at high levels and releases two types of agents that move along adjacent East-step and non-East-step positions. Reading the symbols encountered by these agents, level by level from top to bottom, produces the output word.

\paragraph{Scaffolding map.}
\begin{enumerate}
    \item Group peak positions by their level $\ell_i$.
    \item Initialize $\textit{out}=[\,]$, $\textit{agents}=[\,]$, and set the current level to the maximum level of any peak.
    \item While $|\textit{out}|<2n$:
    \begin{enumerate}
        \item Let $\textit{queue}$ be the union of the current agents and the peak positions at the current level.
        \item Sort $\textit{queue}$ in decreasing order.  For each $i$ in the queue in this order, append $w_i$ to $\textit{out}$.
        \item Update all existing agents simultaneously.  If an agent is at a position $i$ in $R$, move it to $i+1$ when this is a valid position in $R$; otherwise remove it.  If an agent is at a position $i$ not in $R$, move it to $i-1$ when this is a valid position not in $R$; otherwise remove it.
        \item For each peak $j$ at the current level, spawn new agents at $j+1$ if $j+1\in R$ and at $j-1$ if $j-1\notin R$, when valid.
        \item Decrease the current level by $1$.
    \end{enumerate}
    \item Return $\textit{out}$.
\end{enumerate}

We call this the \emph{scaffolding map} because the upward steps build the level structure, while the output is generated by agents traversing this scaffold from peak positions. The name emphasizes the separation between the latent support structure and the subsequent traversal. Although the transformer is trained at a fixed semilength $n$, the resulting \emph{scaffolding map} is defined uniformly for Dyck paths of all semilengths.

\begin{theorem}
\label{thm:scaffolding-zeta}
The scaffolding map is equivalent to the zeta map up to a reversal convention in the labeling. Precisely, denote the scaffolding map by $\varphi$. Define $\mathrm{rev}$ to be flipping the Dyck path with respect to $y= -x$, or reading the Dyck word backward with $0$ and $1$ swapped. Then
\[ \varphi = \mathrm{rev} \circ \zeta . \] 
\end{theorem}

The full proof is provided in \Cref{app:proof}.  The proof strategy is to compare the scaffolding map with the area-sequence description of $\rev\circ\zeta$ level by level. The model-derived algorithm is therefore not presented as a replacement for proof, but as a source of a precise theorem whose verification can be completed by standard combinatorial arguments.

\section{Discussions}
\paragraph{Summary.}
This work uses the zeta map on Dyck paths as a controlled test case for AI-assisted mathematical discovery. By compressing the model to a minimal one-head encoder--decoder transformer, we obtain a learned solution whose internal structure can be inspected directly. The resulting attention, causal ablation, and probing evidence point to a level-based computation, which we translate into the scaffolding map: an explicit peak-centered traversal algorithm for Dyck paths. This illustrates a workflow in which a neural model is not the final mathematical object, but a source of structured evidence that guides the formulation of a new theorem.

\paragraph{Limitations.} The present study has several limitations. First, attention patterns alone are not proof of mechanism \cite{jain2019attentionexplanation, wiegreffe2019attentionexplanation}; this is why we pair them with causal ablations, probes, and an independent combinatorial formulation. Second, the transformer we interpret appears to implement a ``fuzzy'' version of the scaffolding algorithm. This is reflected, for example, in the imperfect level-probing results. The learned mechanism may become cleaner with longer training, as observed in related settings~\citep{nanda2023progress}. More importantly, the scaffolding map was extracted from a particularly simple trained model. Larger models may implement complicated algorithms, and we also observe even small models may admit multiple internal solutions across random seeds~\cite{zhong2023clockpizzastoriesmechanistic,wen2023transformersuninterpretablemyopicmethods}.

\paragraph{Future Directions.} Future work could automate more of this discovery loop and adapt it to larger, less transparent models. Automated circuit-discovery methods and interpretability agents~\citep{conmy2023towards,friedman2023learning,shaham2024multimodal}, together with activation-to-language methods~\citep{pan2026latentqateachingllmsdecode,karvonen2026activationoraclestrainingevaluating}, could help identify recurring internal mechanisms and translate them into candidate symbolic rules. Coupling these tools with theorem provers may further support the transition from model-derived hypotheses to formally verified combinatorial algorithms.

\bibliography{FPSAC2026}

\begin{thebibliography}{30}
\providecommand{\natexlab}[1]{#1}
\providecommand{\url}[1]{\texttt{#1}}
\expandafter\ifx\csname urlstyle\endcsname\relax
  \providecommand{\doi}[1]{doi: #1}\else
  \providecommand{\doi}{doi: \begingroup \urlstyle{rm}\Url}\fi

\bibitem[Alain \& Bengio(2016)Alain and Bengio]{alain2016understanding}
Alain, G. and Bengio, Y.
\newblock Understanding intermediate layers using linear classifier probes.
\newblock \emph{arXiv preprint arXiv:1610.01644}, 2016.

\bibitem[Andrews et~al.(2002)Andrews, Krattenthaler, Orsina, and Papi]{andrews_ad-nilpotent_2002}
Andrews, G., Krattenthaler, C., Orsina, L., and Papi, P.
\newblock ad-{Nilpotent} b-{Ideals} in sl(n) {Having} a {Fixed} {Class} of {Nilpotence}: {Combinatorics} and {Enumeration}.
\newblock \emph{Transactions of the American Mathematical Society}, 354\penalty0 (10):\penalty0 3835--3853, 2002.
\newblock ISSN 0002-9947.
\newblock URL \url{https://www.jstor.org/stable/3072985}.
\newblock Publisher: American Mathematical Society.

\bibitem[Armstrong et~al.(2015)Armstrong, Loehr, and Warrington]{armstrong2015sweep}
Armstrong, D., Loehr, N., and Warrington, G.
\newblock {Sweep maps: A continuous family of sorting algorithms}.
\newblock \emph{Advances in Mathematics}, 284:\penalty0 159--185, 2015.

\bibitem[Babei et~al.(2025)Babei, Charton, Costa, Huang, Lee, Lowry-Duda, Narayanan, and Pozdnyakov]{babei2025learning}
Babei, A., Charton, F., Costa, E., Huang, X., Lee, K.-H., Lowry-Duda, D., Narayanan, A., and Pozdnyakov, A.
\newblock Learning euler factors of elliptic curves, 2025.

\bibitem[B{\'e}rczi et~al.(2026)B{\'e}rczi, Hashemi, and Kl{\"u}ver]{berczi2026flow}
B{\'e}rczi, G., Hashemi, B., and Kl{\"u}ver, J.
\newblock Flow-based extremal mathematical structure discovery.
\newblock \emph{arXiv preprint arXiv:2601.18005}, 2026.

\bibitem[Bills et~al.(2023)Bills, Cammarata, Mossing, Tillman, Gao, Goh, Sutskever, Leike, Wu, and Saunders]{bills2023language}
Bills, S., Cammarata, N., Mossing, D., Tillman, H., Gao, L., Goh, G., Sutskever, I., Leike, J., Wu, J., and Saunders, W.
\newblock Language models can explain neurons in language models.
\newblock \url{https://openaipublic.blob.core.windows.net/neuron-explainer/paper/index.html}, 2023.

\bibitem[Butbaia et~al.(2026)Butbaia, Orland, Huang, Passaro, Fagan, Tarquini, Dao, Eisenbud, Shehper, and Gukov]{Butbaia2026hierarchical}
Butbaia, G., Orland, P., Huang, X., Passaro, D., Fagan, L., Tarquini, M., Dao, H., Eisenbud, D., Shehper, A., and Gukov, S.
\newblock Hierarchical reinforcement learning for sparse-reward search in commutative algebra.
\newblock In \emph{Forty-third International Conference on Machine Learning}, 2026.
\newblock URL \url{https://openreview.net/forum?id=DF6jVG4fG8}.

\bibitem[Charton et~al.(2024)Charton, Ellenberg, Wagner, and Williamson]{charton2024patternboost}
Charton, F., Ellenberg, J., Wagner, A., and Williamson, G.
\newblock {Patternboost: Constructions in mathematics with a little help from ai}, 2024.

\bibitem[Chughtai et~al.(2023)Chughtai, Chan, and Nanda]{chughtai2023toy}
Chughtai, B., Chan, L., and Nanda, N.
\newblock A toy model of universality: Reverse engineering how networks learn group operations.
\newblock In \emph{International Conference on Machine Learning}, pp.\  6243--6267. PMLR, 2023.

\bibitem[Conmy et~al.(2023)Conmy, Mavor-Parker, Lynch, Heimersheim, and Garriga-Alonso]{conmy2023towards}
Conmy, A., Mavor-Parker, A., Lynch, A., Heimersheim, S., and Garriga-Alonso, A.
\newblock Towards automated circuit discovery for mechanistic interpretability.
\newblock \emph{Advances in Neural Information Processing Systems}, 36:\penalty0 16318--16352, 2023.

\bibitem[Fagan et~al.(2026)Fagan, Tarquini, Shehper, Manko, Huang, Gruen, Butbaia, Passaro, and Gukov]{Fagan2026the}
Fagan, L., Tarquini, M., Shehper, A., Manko, M., Huang, X., Gruen, A., Butbaia, G., Passaro, D., and Gukov, S.
\newblock The two-hump problem: Bridging the difficulty gap in mathematical reinforcement learning.
\newblock In \emph{Forty-third International Conference on Machine Learning}, 2026.
\newblock URL \url{https://openreview.net/forum?id=PUQ0rrmuDM}.

\bibitem[Friedman et~al.(2023)Friedman, Wettig, and Chen]{friedman2023learning}
Friedman, D., Wettig, A., and Chen, D.
\newblock Learning transformer programs.
\newblock \emph{Advances in Neural Information Processing Systems}, 36:\penalty0 49044--49067, 2023.

\bibitem[Gukov et~al.(2021)Gukov, Halverson, Ruehle, and Su{\l}kowski]{gukov2021learning}
Gukov, S., Halverson, J., Ruehle, F., and Su{\l}kowski, P.
\newblock Learning to unknot.
\newblock \emph{Machine Learning: Science and Technology}, 2\penalty0 (2):\penalty0 025035, 2021.

\bibitem[Haglund(2003)]{haglund_conjectured_2003}
Haglund, J.
\newblock Conjectured statistics for the \textit{q},\textit{t}-{Catalan} numbers.
\newblock \emph{Advances in Mathematics}, 175\penalty0 (2):\penalty0 319--334, May 2003.
\newblock ISSN 0001-8708.
\newblock \doi{10.1016/S0001-8708(02)00061-0}.
\newblock URL \url{https://www.sciencedirect.com/science/article/pii/S0001870802000610}.

\bibitem[Haiman(2000)]{Haiman2000_qtCatalan}
Haiman, M.
\newblock {The q,t–Catalan Numbers and the Alternating Component of the Diagonal Harmonics}.
\newblock Preprint, University of California, Berkeley, 2000.

\bibitem[Hashemi et~al.(2025)Hashemi, Corominas, and Giacchetto]{hashemi2025can}
Hashemi, B., Corominas, R., and Giacchetto, A.
\newblock Can transformers do enumerative geometry?
\newblock In \emph{International Conference on Learning Representations}, volume 2025, pp.\  70047--70067, 2025.

\bibitem[Jain \& Wallace(2019)Jain and Wallace]{jain2019attentionexplanation}
Jain, S. and Wallace, B.~C.
\newblock Attention is not explanation, 2019.
\newblock URL \url{https://arxiv.org/abs/1902.10186}.

\bibitem[Karvonen et~al.(2026)Karvonen, Chua, Dumas, Fraser-Taliente, Kantamneni, Minder, Ong, Sharma, Wen, Evans, and Marks]{karvonen2026activationoraclestrainingevaluating}
Karvonen, A., Chua, J., Dumas, C., Fraser-Taliente, K., Kantamneni, S., Minder, J., Ong, E., Sharma, A.~S., Wen, D., Evans, O., and Marks, S.
\newblock Activation oracles: Training and evaluating llms as general-purpose activation explainers, 2026.
\newblock URL \url{https://arxiv.org/abs/2512.15674}.

\bibitem[Nanda et~al.(2023)Nanda, Chan, Lieberum, Smith, and Steinhardt]{nanda2023progress}
Nanda, N., Chan, L., Lieberum, T., Smith, J., and Steinhardt, J.
\newblock Progress measures for grokking via mechanistic interpretability, 2023.

\bibitem[Novikov et~al.(2025)Novikov, V{\~u}, Eisenberger, Dupont, Huang, Wagner, Shirobokov, Kozlovskii, Ruiz, Mehrabian, et~al.]{novikov2025alphaevolve}
Novikov, A., V{\~u}, N., Eisenberger, M., Dupont, E., Huang, P.-S., Wagner, A.~Z., Shirobokov, S., Kozlovskii, B., Ruiz, F.~J., Mehrabian, A., et~al.
\newblock Alphaevolve: A coding agent for scientific and algorithmic discovery.
\newblock \emph{arXiv preprint arXiv:2506.13131}, 2025.

\bibitem[Pan et~al.(2026)Pan, Chen, and Steinhardt]{pan2026latentqateachingllmsdecode}
Pan, A., Chen, L., and Steinhardt, J.
\newblock Latentqa: Teaching llms to decode activations into natural language, 2026.
\newblock URL \url{https://arxiv.org/abs/2412.08686}.

\bibitem[Shaham et~al.(2024)Shaham, Schwettmann, Wang, Rajaram, Hernandez, Andreas, and Torralba]{shaham2024multimodal}
Shaham, T.~R., Schwettmann, S., Wang, F., Rajaram, A., Hernandez, E., Andreas, J., and Torralba, A.
\newblock A multimodal automated interpretability agent.
\newblock In \emph{Forty-first International Conference on Machine Learning}, 2024.

\bibitem[{The Sage Developers}(2026)]{sagemath}
{The Sage Developers}.
\newblock \emph{{S}ageMath, the {S}age {M}athematics {S}oftware {S}ystem ({V}ersion 9.5)}, 2026.
\newblock {\tt https://www.sagemath.org}.

\bibitem[Thomas \& Williams(2018)Thomas and Williams]{thomas2018sweeping}
Thomas, H. and Williams, N.
\newblock Sweeping up zeta.
\newblock \emph{Selecta Mathematica}, 24\penalty0 (3):\penalty0 2003--2034, 2018.

\bibitem[Vaswani et~al.(2017)Vaswani, Shazeer, Parmar, Uszkoreit, Jones, Gomez, Kaiser, and Polosukhin]{vaswani2017attention}
Vaswani, A., Shazeer, N., Parmar, N., Uszkoreit, J., Jones, L., Gomez, A., Kaiser, {\L}., and Polosukhin, I.
\newblock Attention is all you need.
\newblock \emph{Advances in neural information processing systems}, 30, 2017.

\bibitem[Wagner(2021)]{wagner2021constructionscombinatoricsneuralnetworks}
Wagner, A.~Z.
\newblock Constructions in combinatorics via neural networks, 2021.
\newblock URL \url{https://arxiv.org/abs/2104.14516}.

\bibitem[Wen et~al.(2023)Wen, Li, Liu, and Risteski]{wen2023transformersuninterpretablemyopicmethods}
Wen, K., Li, Y., Liu, B., and Risteski, A.
\newblock Transformers are uninterpretable with myopic methods: a case study with bounded dyck grammars, 2023.
\newblock URL \url{https://arxiv.org/abs/2312.01429}.

\bibitem[Wennberg \& Henter(2024)Wennberg and Henter]{wennberg2024learned}
Wennberg, U. and Henter, G.
\newblock Learned transformer position embeddings have a low-dimensional structure.
\newblock In \emph{Proceedings of the 9th workshop on representation learning for NLP (RepL4NLP-2024)}, pp.\  237--244, 2024.

\bibitem[Wiegreffe \& Pinter(2019)Wiegreffe and Pinter]{wiegreffe2019attentionexplanation}
Wiegreffe, S. and Pinter, Y.
\newblock Attention is not not explanation, 2019.
\newblock URL \url{https://arxiv.org/abs/1908.04626}.

\bibitem[Zhong et~al.(2023)Zhong, Liu, Tegmark, and Andreas]{zhong2023clockpizzastoriesmechanistic}
Zhong, Z., Liu, Z., Tegmark, M., and Andreas, J.
\newblock The clock and the pizza: Two stories in mechanistic explanation of neural networks, 2023.

\end{thebibliography}
\bibliographystyle{icml2026}

\appendix
\onecolumn

\section{$q,t$-Catalan Numbers and the Zeta Map}
\label{app:qt_cat}
\subsection{$q,t$-Catalan Numbers}

We start this section by defining the main objects of interest, Dyck paths.

\begin{definition}
    A Dyck path of semilength $n$ is a length $2n$ binary sequence $w = (w_i)_{i=1,\ldots,2n}$ with the property that each initial segment $(w_i)_{i=1,\ldots,k; \ k \leq 2n}$ of $w$ contains at least as many $1$'s as $0$'s. We denote the set of all Dyck paths of semilength $n$ by $Dyck(n)$.
\end{definition}

\begin{rmk}
    Equivalently, a Dyck path of semilength $n$ is a lattice path from $(0,0)$ to $(n,n)$ consisting of North $(0,1)$ and East $(1,0)$ steps that stays on or above the line $y=x$. Under this equivalence, a $1$ in the binary representation corresponds to a North step, and a $0$ corresponds to an East step. We will use $w$ to refer to the binary sequence and lattice path representation of a Dyck path interchangeably.
\end{rmk}

It is well known that the number of Dyck paths of semilength $n$ is the $n$th Catalan number \[ C_n = \frac{1}{n+1} {2n \choose n}. \]

There is another equivalent formalization of Dyck paths in terms of the number of complete boxes between the lattice path representation of a Dyck path and the diagonal $y=x$ in each row, from bottom to top, which is convenient for calculating the statistics in Theorem 1. 

\begin{definition}
    The area sequence of a Dyck path $w$ of semilength $n$ is the sequence $(a_i)_{i=1,\ldots,n}$ where $a_i$ counts the number of complete boxes in row $i$ between the Dyck path and the diagonal $y=x$, where the rows are enumerated from bottom to top.
\end{definition}

Every Dyck path $w$ uniquely determines an area sequence $(a_i)$ and vice versa. An example of the lattice path, binary sequence, and area sequence representations of a Dyck path is given in Example 1.

We now define the three statistics which appear in \Cref{eq:two-formulas}.

\begin{definition}
    Let $w = (w_i)$ be a Dyck path of semilength $n$ and $(a_i)$ be its area sequence. The area of $w$ is the nonnegative integer \[ area(w) = \sum_{i=1}^{n} a_i. \] The dinv (or diagonal inversion) of $w$ is the nonnegative integer \[ dinv(w) = \#\{ (i,j) \ | \ i < j,  a_i = a_j\} + \#\{ (i,j) \ | \ i < j,  a_i = a_j + 1 \}.  \] 
\end{definition}

\begin{definition}
    Let $w = (w_i)$ be a Dyck path of semilength $n$. The bounce of $w$ is the nonnegative integer calculated in the following way:
    \begin{enumerate}
                \item Shoot a billiard from $(0,0)$ heading North
                \item When you hit the start of an East step, turn East and travel until you hit the diagonal
                \item Turn North and repeat until you reach $(n,n)$.
                \item Record the places you hit the diagonal $(0,0), (j_1,j_1),\ldots, (j_k,j_k), (n,n)$.
            \end{enumerate}
    Then \[ bounce(w) = \sum_{i=1}^k (n - j_i). \]
\end{definition}

\begin{definition}
    The $n^{th}$ $q,t$-Catalan number is \[ C_n(q,t) = \sum_{w \in Dyck(n)} q^{area(w)}t^{bounce(w)} = \sum_{w \in Dyck(n)} q^{dinv(w)}t^{area(w)} \in \mathbb{Z}_{>0}[q,t]. \]
\end{definition}

\subsection{Haglund's Zeta Map}
\label{sec:exts-alg}

The zeta map uses the area sequence/bounce path \cite{andrews_ad-nilpotent_2002,haglund_conjectured_2003}, and can be described by the following algorithm:

\begin{enumerate}
    \item Compute the area sequence $a_1, a_2, \ldots, a_n$ of $w$ (row lengths from bottom).
    \item Set $b = \text{max}\{ a_i \}+1$.
    \item For each $k=0, 1, \ldots , b$,   scan the sequence $a_1, a_2, \ldots, a_n$ from left to right and record:
    \begin{enumerate}
        \item a \(0\) for each occurrence of \(k-1\),
        \item a \(1\) for each occurrence of \(k\).
    \end{enumerate}
Since the area sequence contains no entries equal to \(-1\) or \(b\), this procedure starts with  a \(1\) when \(k=0\) and ends with a \(0\) when \(k=b\).\end{enumerate}

Below is an example of the image of the zeta map of a Dyck path of semilength 8.
\begin{ex} \label{ex-running}
    Consider the Dyck path $w$ below.
    \[\begin{tikzpicture}[scale=0.5]
      \dyckpath{0,0}{8}{1,1,1,0,1,0,1,1,0,0,0,1,1,0,0,0}{dyck1}{};
    \end{tikzpicture}\]
The area sequence is $(0,1,2,2,2,3,1,2)$. Scanning from left to right for $-1$ and $0$, we find no occurrence of $-1$, and the single occurrence of $0$ produces $1$ in the image under the zeta map.
Next, scanning for $0$ and $1$, we obtain the subsequence $0,1,1$, which yields $0,1,1$ in the image.
Scanning for $1$ and $2$ gives $1,2,2,2,1,2$, producing $0,1,1,1,0,1$.
Similarly, scanning for $2$ and $3$ yields $2,2,2,3,2$, which produces $0,0,0,1,0$.
Finally, scanning for $3$ and $4$ gives $3$, producing $0$.

    \[\begin{tikzpicture}[scale=0.3]
      \dyckpath{0,0}{8}{1}{dyck1}{};
      \node at (9.5,3.5) {$\longrightarrow$};
      \dyckpath{11,0}{8}{1,0,1,1}{dyck2}{};
      \node at (20.5,3.5) {$\longrightarrow$};
      \dyckpath{22,0}{8}{1,0,1,1,0,1,1,1,0,1}{dyck3}{};
    \node at (31.5,3.5) {$\longrightarrow$};
      \dyckpath{33,0}{8}{1,0,1,1,0,1,1,1,0,1,0,0,0,1,0}{dyck4}{};
      \node at (42.5,3.5) {$\longrightarrow$};
      \dyckpath{44,0}{8}{1,0,1,1,0,1,1,1,0,1,0,0,0,1,0,0}{dyck5}{};
    \end{tikzpicture}\]
\end{ex}

\section{Architecture of Minimal Dyck Transformer}
\label{app:minimal_Dyck}

We use a deliberately small encoder--decoder transformer. The goal is to keep architectural capacity minimal while still permitting explicit inspection of attention patterns.

\paragraph{Model specification.}
The model is a standard encoder--decoder transformer without dropout:
\begin{itemize}
    \item Embedding/model width: $d_{\text{model}} = 128$.
    \item Feed-forward width: $d_{\text{ff}} = 256$ with GELU nonlinearity.
    \item Layers: $1$ encoder block and $1$ decoder block.
    \item Attention heads: $1$.
    \item Dropout: $0$.
    \item Maximum length: positional embeddings defined up to $\texttt{max\_len}=128$.
\end{itemize}
It has approximately $3.66\times 10^5$ trainable parameters.

\paragraph{Token and positional embeddings.}
For a token sequence $(x_1,\dots,x_L)$, the encoder input is
\[
\mathbf{H} = \mathrm{Embed}_{src}(x_{1:L}) + \mathrm{PosEmbed}(1{:}L),
\]
and the decoder input is defined similarly, using a distinct target embedding table $\mathrm{Embed}_{tgt}$.
Positional encodings are \emph{learned}, i.e., given by an embedding lookup indexed by position; sinusoidal encoding is disabled in this configuration.

\paragraph{Attention mechanism.}
All attention sublayers use scaled dot-product attention. For queries $Q$, keys $K$, and values $V$,
$$
\mathrm{Attn}(Q,K,V) = \mathrm{softmax}\!\left(\frac{QK^\top}{\sqrt{d_{\text{head}}}} \right)V.
$$
Decoder \emph{self-attention} is causal: when no mask is explicitly provided, an upper-triangular causal mask is applied to prevent attention to future target positions. Encoder self-attention and decoder cross-attention are non-causal.

\paragraph{Transformer blocks (post-norm).}
Each encoder block applies self-attention and a position-wise feed-forward network with residual connections and \emph{post-layer normalization}:
\[
\mathbf{H} \leftarrow \mathrm{LN}\!\big(\mathbf{H} + \mathrm{SelfAttn}(\mathbf{H})\big), \quad
\mathbf{H} \leftarrow \mathrm{LN}\!\big(\mathbf{H} + \mathrm{FFN}(\mathbf{H})\big).
\]
Each decoder block applies (i) causal self-attention, (ii) encoder--decoder cross-attention, and (iii) the same FFN, each followed by residual addition and post-norm:
\[
\mathbf{Z} \leftarrow \mathrm{LN}\!\big(\mathbf{Z} + \mathrm{SelfAttn}_{\text{causal}}(\mathbf{Z})\big),\;
\mathbf{Z} \leftarrow \mathrm{LN}\!\big(\mathbf{Z} + \mathrm{CrossAttn}(\mathbf{Z}, \mathbf{H}_{enc})\big),\;
\mathbf{Z} \leftarrow \mathrm{LN}\!\big(\mathbf{Z} + \mathrm{FFN}(\mathbf{Z})\big).
\]
The FFN is
\[
\mathrm{FFN}(\mathbf{x}) = W_2\,\mathrm{GELU}(W_1\mathbf{x}+b_1)+b_2,
\]
with $W_1\in\mathbb{R}^{128\times256}$ and $W_2\in\mathbb{R}^{256\times128}$.

\paragraph{Output head.}
Decoder hidden states are mapped to next-token logits by a linear projection $\mathbb{R}^{128}\rightarrow\mathbb{R}^{4}$ applied at each position. No weight tying is used between the embeddings and the output projection.

\newpage
\section{Other Embedding And Attention Visualizations}
\label{app:other_visuals}
In this section, we present some visualizations that were helpful when investigating the internal mechanism of the transformer.

\paragraph{Additional Cross-Attention Examples.}
\label{app:attention}
We include additional examples illustrating the three recurring attention patterns, showing that they are common phenomena rather than hand-picked examples: highest-level selection, systematic avoidance of many positions with $w_i=1$, and triangular sequential shift.

\begin{figure}[H]
    \centering
    \includegraphics[width=\linewidth]{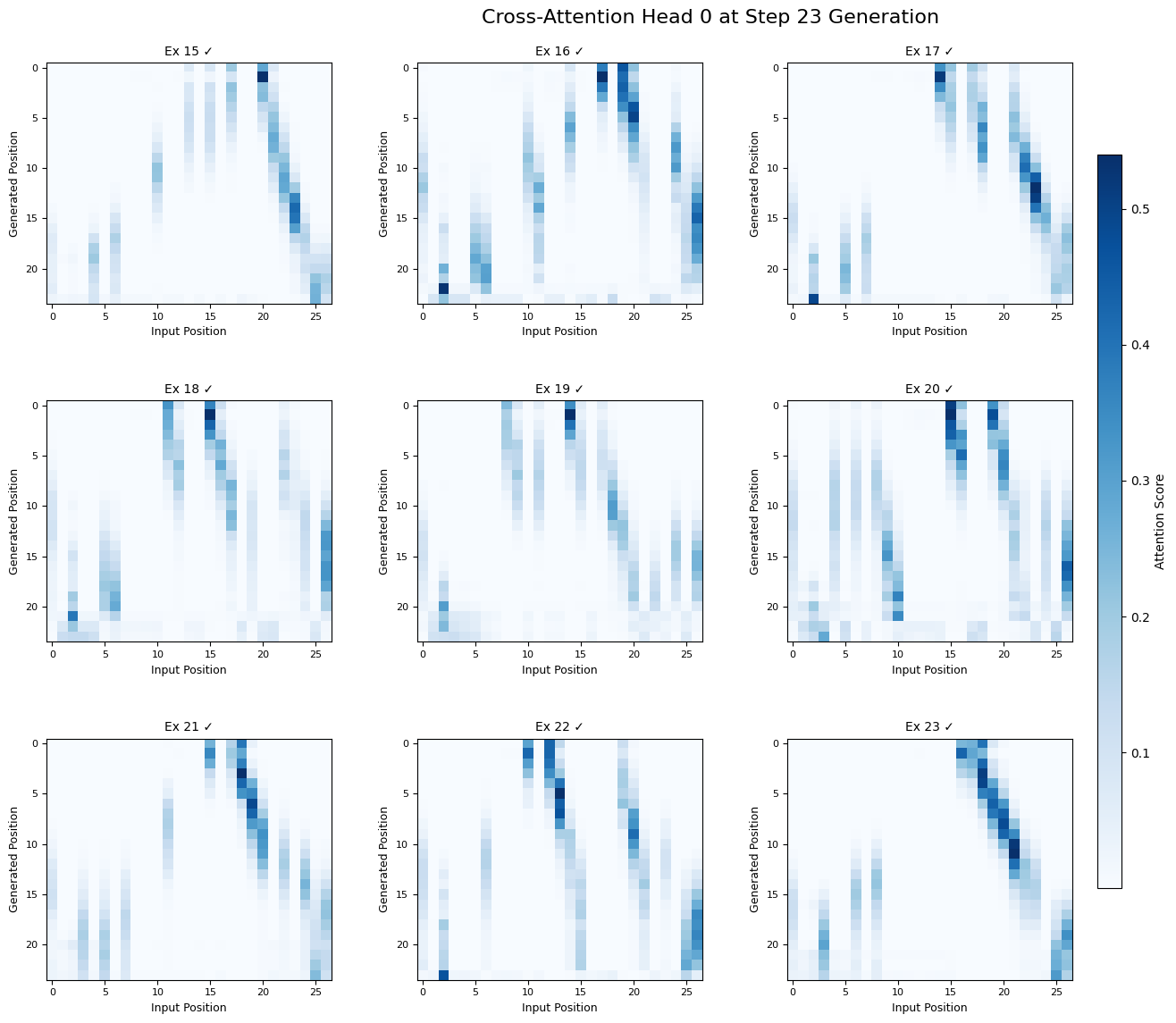}
    \caption{Cross-attention matrices at a fixed output step $23$ generation for a batch of random samples, illustrating that the triangular shift pattern occurs consistently across examples.}
    \label{fig:xattn_batch}
\end{figure}

\newpage
\begin{figure}[H]
    \centering
    \includegraphics[height=0.92\textheight,keepaspectratio]{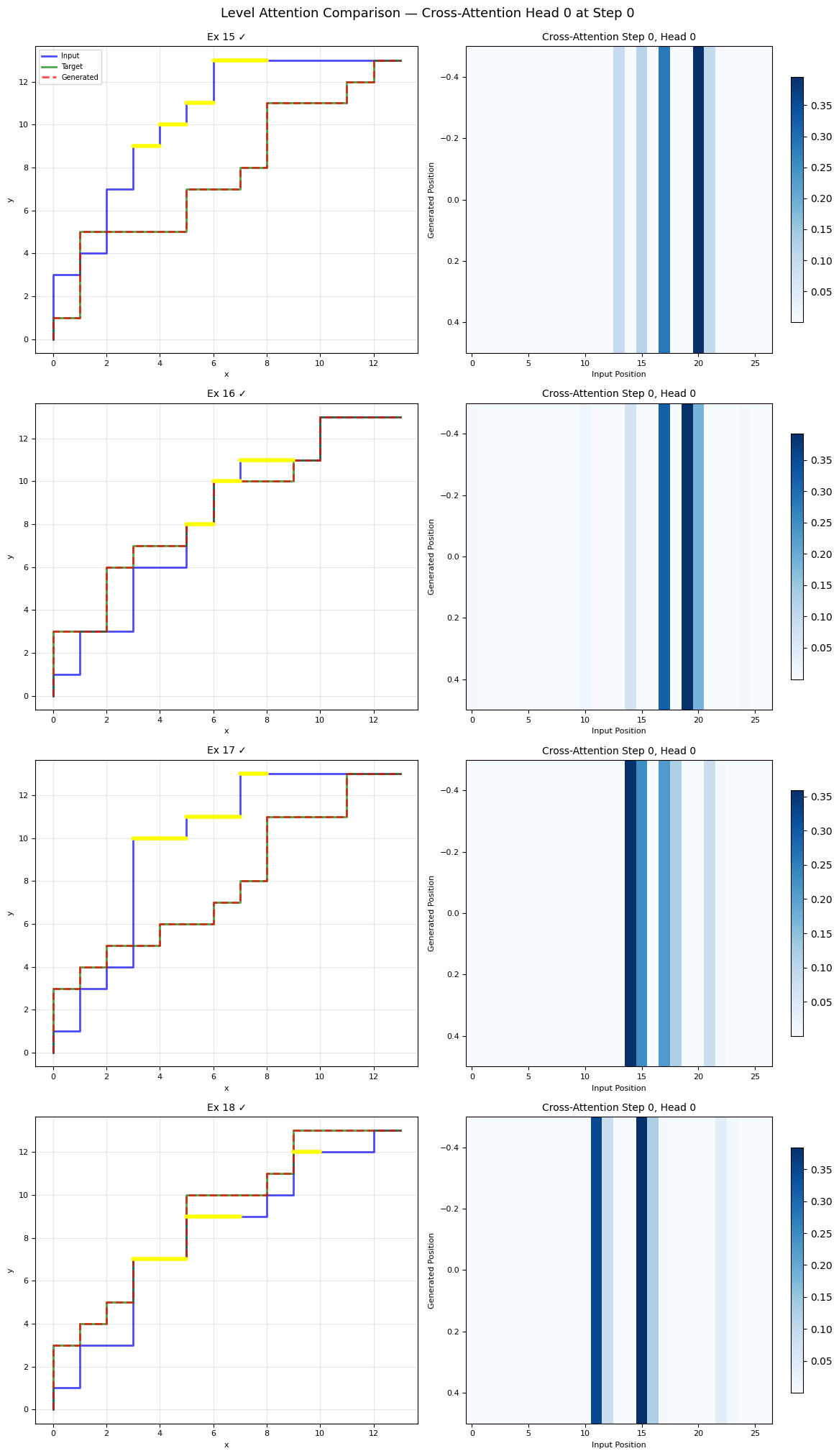}
    \caption{Level-attention comparison across examples. Decoder cross-attention at the first output step consistently concentrates on high-level input positions.}
    \label{fig:level_attention}
\end{figure}

\newpage
\paragraph{Encoder self-attention.}
Beyond decoder cross-attention, we also inspect encoder self-attention to understand how the model organizes input positions before decoding. The resulting attention maps show recurring block-like structure partly due to the binary nature of Dyck paths. 

\begin{figure}[H]
    \centering
    \includegraphics[width=\linewidth]{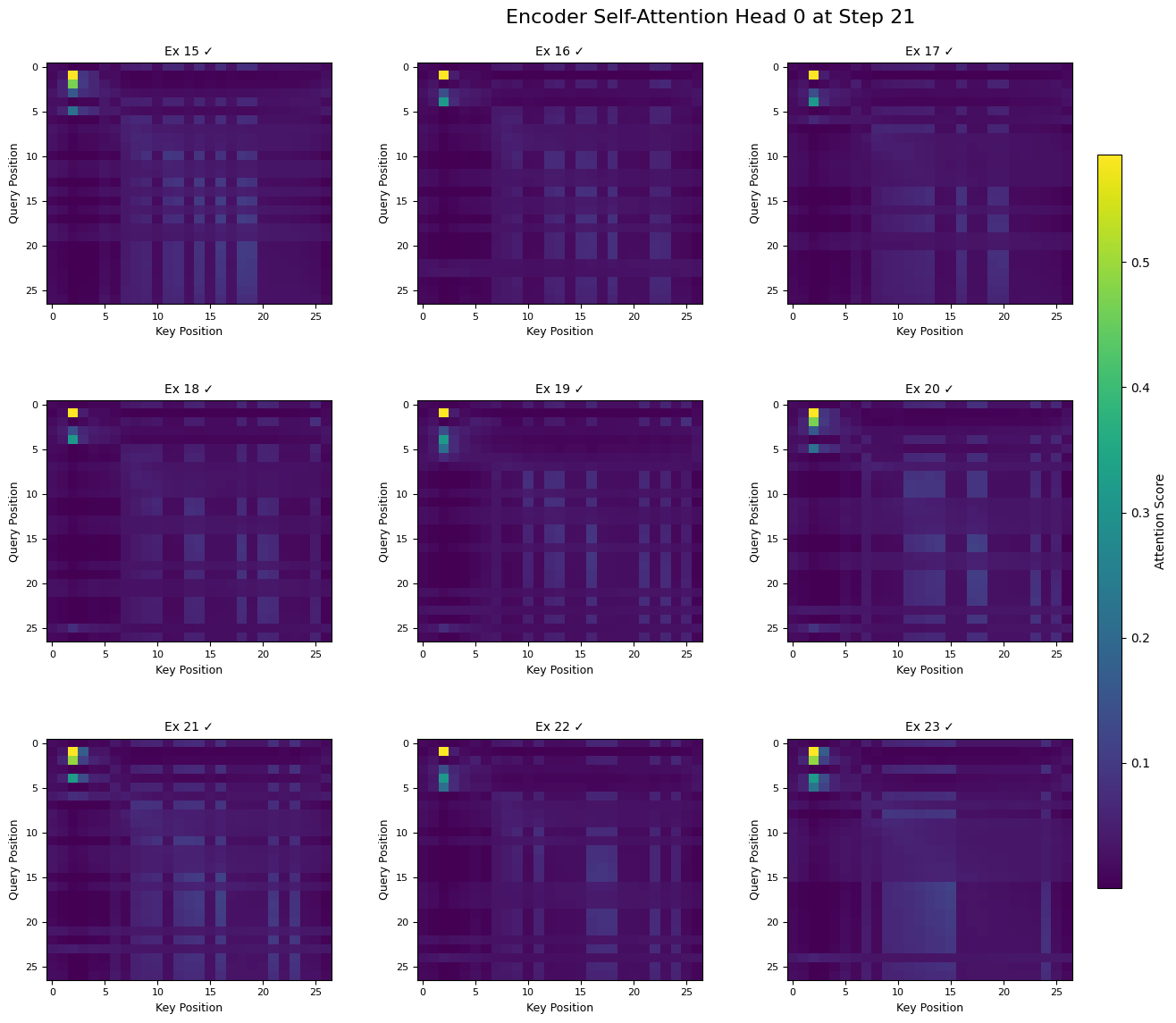}
    \caption{Encoder self-attention matrices across a batch of examples. The recurring block-like structure reflects the binary organization of Dyck words and suggests that the encoder forms structured position-level representations.}
    \label{fig:encoder_attn}
\end{figure}

\newpage
\paragraph{Learned positional embeddings.}
We further visualize the learned encoder and decoder positional embeddings using PCA. These plots show that the model learns nontrivial positional geometry: for the encoder embeddings, the first principal component separates early positions from the main cluster, while the second principal component largely follows the ordering of input positions. This ordered geometry may help explain the block-like structure observed in the encoder attention maps. Since nearby positions have similar positional embeddings, their position-dependent contributions to attention are similar; consequently, much of the remaining variation in the attention pattern is driven by the binary token value, \(0\) or \(1\). This combination of smooth positional variation and token-level separation naturally produces block-like attention structure.

\begin{figure}[H]
    \centering
    \includegraphics[width=\linewidth]{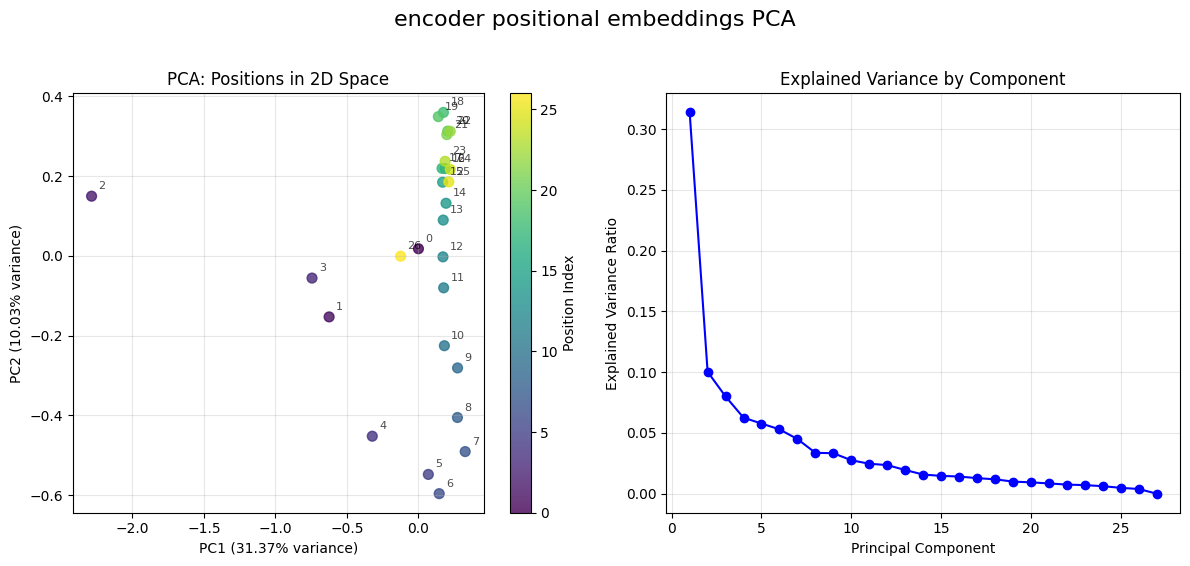}
    \caption{PCA of the learned encoder positional embeddings. The second principal component largely follows the ordering of input positions, while the first component separates some early positions from the main cluster.}
    \label{fig:encoder_pos_emb}
\end{figure}

\begin{figure}[H]
    \centering
    \includegraphics[width=\linewidth]{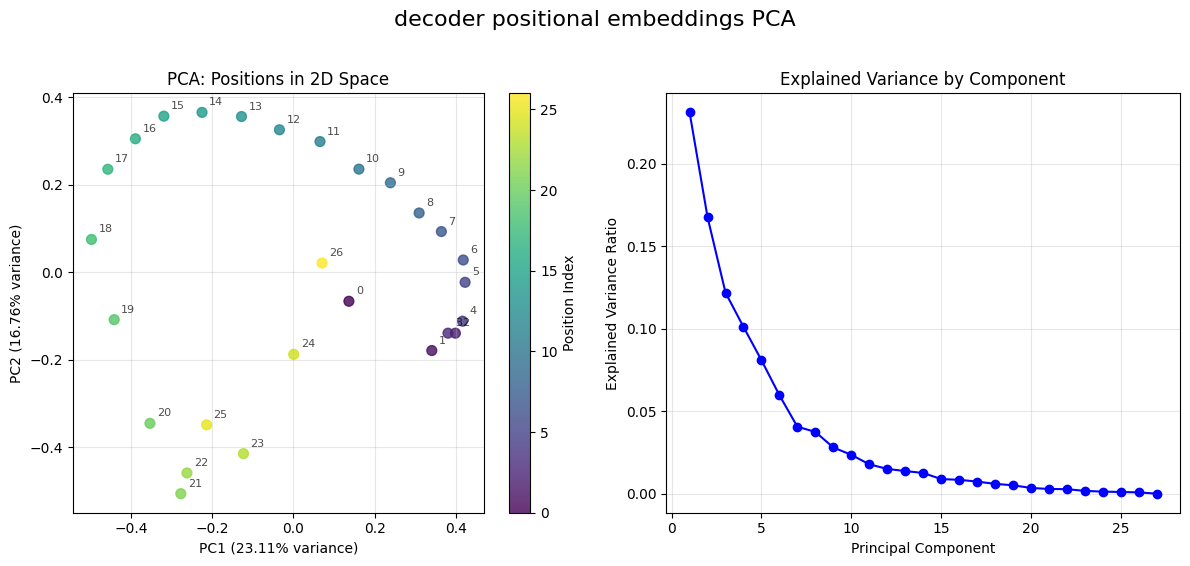}
    \caption{PCA of the learned decoder positional embeddings. Decoder positions form a more regular spiral trajectory in the first two principal components~\citep{wennberg2024learned,vaswani2017attention}.}
    \label{fig:decoder_pos_emb}
\end{figure}

\newpage
\section{Proof of the Scaffolding--Zeta Equivalence}
\label{app:proof}

To begin with, we clarify conventions and fix notations.

Our scaffolding map, denoted by \( \varphi\), coincides with \texttt{area\_dinv\_to\_bounce\_area\_map()} in \textsc{SageMath}, which we used to generate our datasets. 
However, it does not agree with Haglund's zeta map, denoted by \(\zeta\), due to a difference in conventions: the bounce path is defined in the reverse direction in \textsc{SageMath} compared to Haglund's formulation.
    In this section, we prove  \[ \varphi = \mathrm{rev} \circ \zeta , \]  where $\mathrm{rev}$ is reading the Dyck word backward with $0$ and $1$ swapped.

\bigskip

Let $a(w)= (a_1, a_2, \dots , a_n)$ be the area sequence of a Dyck path $w$, and set $b = \text{max}\{ a_i \}+1$. Recall the algorithm for $\zeta$ described at the beginning of Section \ref{sec:exts-alg}.

{\it  
 For each $k=0, 1, \ldots , b$,   scan the sequence $a_1, a_2, \ldots, a_n$ from left to right and record:
    \begin{enumerate}
        \item a \(0\) for each occurrence of \(k-1\),
        \item a \(1\) for each occurrence of \(k\).
\end{enumerate} }
\noindent Accordingly, an algorithm for $\mathrm{rev} \circ \zeta$ is given by the following. 

{\it    
    For each $k=b, b-1, \ldots , 0$,   scan the sequence $a_1, a_2, \ldots, a_n$ from right to left and record:
    \begin{enumerate}
        \item a \(0\) for each occurrence of \(k\),
        \item a \(1\) for each occurrence of \(k-1\).
\end{enumerate}
}
From the algorithm of $\mathrm{rev} \circ \zeta$, for each $k=b, b-1, \ldots , 0$, we obtain a subsequence $P_k$ of $\tilde a(w)$ consisting of $k$ and $k-1$, where $\tilde a(w) = (a_n, a_{n-1}, \dots , a_1)$. (Recall that we scan $a(w)$ from right to left in the algorithm of $\mathrm{rev} \circ \zeta$.) 
For example, the Dyck path in Example \ref{ex-running} (see \Cref{fig:example_with_xo} below) yields 
\[  P_4= (3), \quad P_3= (2,3,2,2,2), \quad P_2= (2,1,2,2,2,1), \quad P_1 = (1,1,0), \quad P_0 =(0).  
\]
These subsequences completely determine the image of $\mathrm{rev} \circ \zeta$. 

On the other hand, in the algorithm of the scaffolding map $\varphi$ described in Section \ref{sec:scaffolding}, each $w_i$ of $w=(w_i)$ is associated with level $\ell_i$, $1 \le i \le 2n$. Define \[ b_i = \begin{cases} \ell_i -1 & \text{if } w_i =1, \\ \ell_i  & \text{if } w_i=0 . \end{cases} \] 
For each $1 \le i \le 2n$, we associate position $i$  with $a_r$ for some $r$ in $a(w) = (a_1, \dots , a_n)$ in the following way; in particular, we will have $b_i =a_r$. 

If $w_i=1$ and position $i$ belongs to the $r^{\mathrm{th}}$ row, then we associate position $i$ with $a_r$ and have $b_i= \ell_i-1 = a_r$.    
If $w_i=0$, then we slide with slope $1$ down to the south-west direction until we hit a position, say $j$, with $w_j=1$. Suppose that position $j$ belongs to the $r^{\mathrm{th}}$ row. Then we associate position $i$ with $a_r$ and obtain $b_i = \ell_i=a_r$ from the symmetry of rows and columns, noting that $b_i$ is equal to the length of the column that contains position $i$. For example, in Example \ref{ex-running}, position 10, marked with $\times$, is associated with $a_5=2$, and position 11, marked with \raisebox{- 1pt}{\scalebox{1.4}{$\circ$}}, is associated with $a_2=1$. 

\begin{figure}[H]
    \centering
    \begin{tikzpicture}[scale=0.5]
        \dyckpath{0,0}{8}{1,1,1,0,1,0,1,1,0,0,0,1,1,0,0,0}{dyck1}{};
        \node at (3.5, 6) {$\times$};
        \draw (4.5, 6) circle (6pt);
        \draw[->] (3.5,5.8) -- (2.2, 4.5);
        \draw[->] (4.5, 5.6) -- (0.2,1.3);
    \end{tikzpicture}
    \caption{Dyck path with marked positions used in the proof.}
    \label{fig:example_with_xo}
\end{figure}
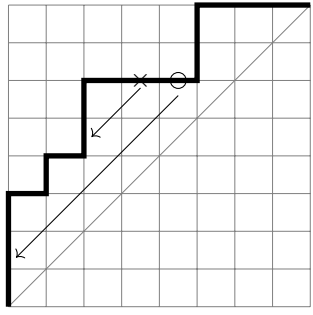

In step (3)(a) of the scaffolding algorithm, the queues $Q_b, Q_{b-1}, \dots, Q_0$ are formed, where $b =\max \{\ell_i \}= \max\{a_i\} +1$ as before. For $k=b,b-1, \dots , 0$, each $Q_k$ consists of positions $i_1, i_2, \dots $. Replace them with the corresponding $b_{i_1}, b_{i_2}, \dots $, and denote the resulting sequence by $Q'_k$. For example, in Example \ref{ex-running}, we have
\begin{eqnarray*}  & Q_4= (8), \quad Q_3= (13,9,7, 5,3), \quad Q_2= (14, 12, 10, 6, 4,2), \quad Q_1 = (15, 11, 1), \quad Q_0 =(16), \\  & Q'_4= (3), \quad Q'_3= (2,3,2,2,2), \quad Q'_2=  (2,1,2,2,2,1), \quad Q'_1 = (1,1,0), \quad Q'_0 =(0) .  
\end{eqnarray*}
The sequences $Q_k$, or the sequences $Q'_k$ determine the image of the scaffolding map $\phi$. 

We now claim that \(P_k = Q'_k\) for \(k = b, b-1, \dots, 0\). Indeed, in each \(P_k\), the occurrences of \(k-1\) correspond to entries \(b_i = k-1\) in \(Q'_k\), arising from peaks or positions \(i \notin R\) in \(Q_k\);  the occurrences of \(k\) correspond to positions \(i \in R\), where \(b_i = k = a_r\), and the \(r^{\mathrm{th}}\) row is associated with \(i\) as described above in the penultimate paragraph.

Since $P_k$ and $Q'_k$ determine the images of $\mathrm{rev}\circ \zeta$ and $\phi$, respectively, the claim implies that 
$ \mathrm{rev} \circ \zeta = \phi $, as desired.

\end{document}